\documentclass[sigconf]{acmart}
\usepackage[ruled,vlined,linesnumbered]{algorithm2e}

\usepackage{amsmath,amssymb}

\usepackage{booktabs}
\usepackage{array}
\usepackage{multirow}
\usepackage{multicol}
\usepackage{rotating}
\usepackage{adjustbox}
\usepackage{arydshln}

\usepackage{graphicx}
\usepackage{subcaption}
\usepackage{tikz}

\usepackage{xcolor}
\usepackage{microtype}

\usepackage{pifont}

\usepackage[table]{xcolor} % 必须引入 xcolor 包
\usepackage{geometry}
\usepackage[absolute,overlay]{textpos}

\usepackage{xurl}
\AtBeginDocument{%
  }

\acmDOI{} % 清空DOI（避免模板自动加载协议图标）
\setcopyright{none}
\renewcommand\footnotetextcopyrightpermission[1]{}
\begin{document}

\begin{textblock*}{\paperwidth}(1.5cm, 1.2cm)
\includegraphics[height=1.5cm]{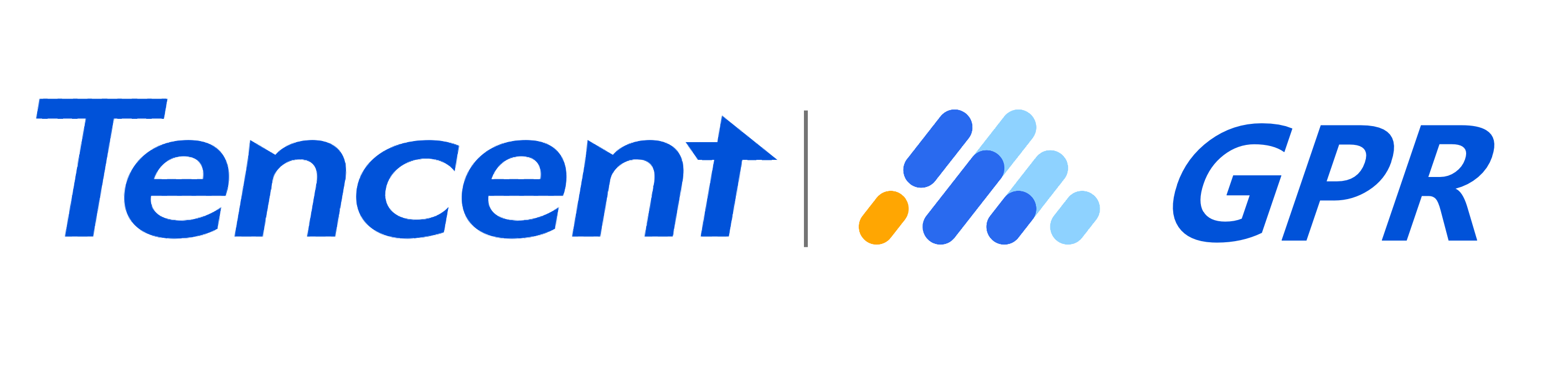} % 请确保目录下有 logo.png
\end{textblock*}

\setlength{\textfloatsep}{2pt}   % floats at top/bottom vs text
\setlength{\intextsep}{2pt}      % floats in the middle of text
\setlength{\abovecaptionskip}{0pt}
\setlength{\belowcaptionskip}{0pt}

\title{Spend Search Where It Pays: Value-Guided Structured Sampling and Optimization for Generative Recommendation}
%{Spend Compute Where It Matters: Value-Guided Efficient Decoding for Generative Recommendation}
%Value-Guided Budgeted Tree Search  for Generative Recommendation
% \author{Anonymous}
% \authornote{Anonymous.}
% \email{Anonymous@Anonymous.com}
% \orcid{Anonymous}
% \affiliation{%
%   \institution{Anonymous}
%   \city{Anonymous}
%   \state{Anonymous}
%   \country{Anonymous}
% }

\author{Jie Jiang$^{*}$, Yangru Huang$^{*}$, Zeyu Wang, Changping Wang, Yuling Xiong, Jun Zhang$^{\dagger}$, Huan Yu}
\affiliation{
  \institution{Tencent Inc., China}
%   \city{Shenzhen}
  \country{}}
\email{{zeus, yarayrhuang, peterzywang, terracewang, whitnyxiong, neoxzhang,huanyu}@tencent.com}

\renewcommand{\shortauthors}{Trovato et al.}

\begin{abstract}

Generative recommendation via autoregressive models has unified retrieval and ranking into a single conditional generation framework. However, fine-tuning these models with Reinforcement Learning (RL) often suffers from a fundamental \textbf{probability-reward mismatch}. Conventional likelihood-dominated decoding (e.g., beam search) exhibits a myopic bias toward locally probable prefixes, which causes two critical failures: (1) \textit{insufficient exploration}, where high-reward items in low-probability branches are prematurely pruned and rarely sampled, and (2) \textit{advantage compression}, where trajectories sharing high-probability prefixes receive highly correlated rewards with low within-group variance, yielding a weak comparative signal for RL.
To address these challenges, we propose \textbf{V-STAR}, a \textbf{V}alue-guided \textbf{S}ampling and \textbf{T}ree-structured \textbf{A}dvantage \textbf{R}einforcement framework. V-STAR forms a self-evolving loop via two {synergistic} components. First, a \textbf{Value-Guided Efficient Decoding (VED)} is developed to identify decisive nodes and selectively deepen high-potential prefixes. This improves exploration efficiency without exhaustive tree search. Second, we propose \textbf{Sibling-GRPO}, which exploits the induced tree topology to compute sibling-relative advantages and concentrates learning signals on decisive branching decisions. Extensive experiments on both offline and online datasets demonstrate that V-STAR outperforms state-of-the-art baselines, delivering superior accuracy and candidate-set diversity under strict latency constraints.

\end{abstract}





% \renewcommand{\thefootnote}{\fnsymbol{footnote}}
% \footnotetext[2]{Corresponding author (neoxzhang@tencent.com).}
% \renewcommand{\thefootnote}{\arabic{footnote}}

\maketitle

\renewcommand{\thefootnote}{\fnsymbol{footnote}}
\footnotetext[1]{Equal contribution: Jie Jiang and Yangru Huang.}
\footnotetext[2]{Corresponding author: Jun Zhang.}
\renewcommand{\thefootnote}{\arabic{footnote}}

\section{Introduction}
Recommender systems are shifting from retrieve-and-rank to end-to-end {generative recommendation (GR)}~\cite{rajput2023recommender, zhai2024actions, li2023gpt4rec}, which has been deployed in e-commerce and short-video platforms~\cite{rajput2023recommender, zhai2024actions}. By reformulating item identifiers into a hierarchical search space of Semantic IDs (SIDs), large-scale candidate generation has become computationally practical with autoregressive language models~\cite{deng2025onerec, wei2025oneloc, yang2025sparse, wang2024eager}. To further bridge the gap between generation likelihood and real-world utility, these GR models are usually fine-tuned with {reinforcement learning (RL)}~\cite{chen2024softmax, chen2025onesearch, sharma2024optimizing} techniques, e.g., on-policy GRPO-style groupwise updates~\cite{deng2025onerec, zhou2025onerec, xing2025reg4rec, hong2025generative, kong2025minionerec}. However, such a training mechanism is bottlenecked by a fundamental probability-reward misalignment. During training, candidate items are typically sampled by probability-driven decoding strategies (e.g., beam search). In contrast, the recommendation objective aims to maximize the expected reward of the full generated SIDs.

\begin{figure}[t]
    \centering
    \includegraphics[width=\linewidth]{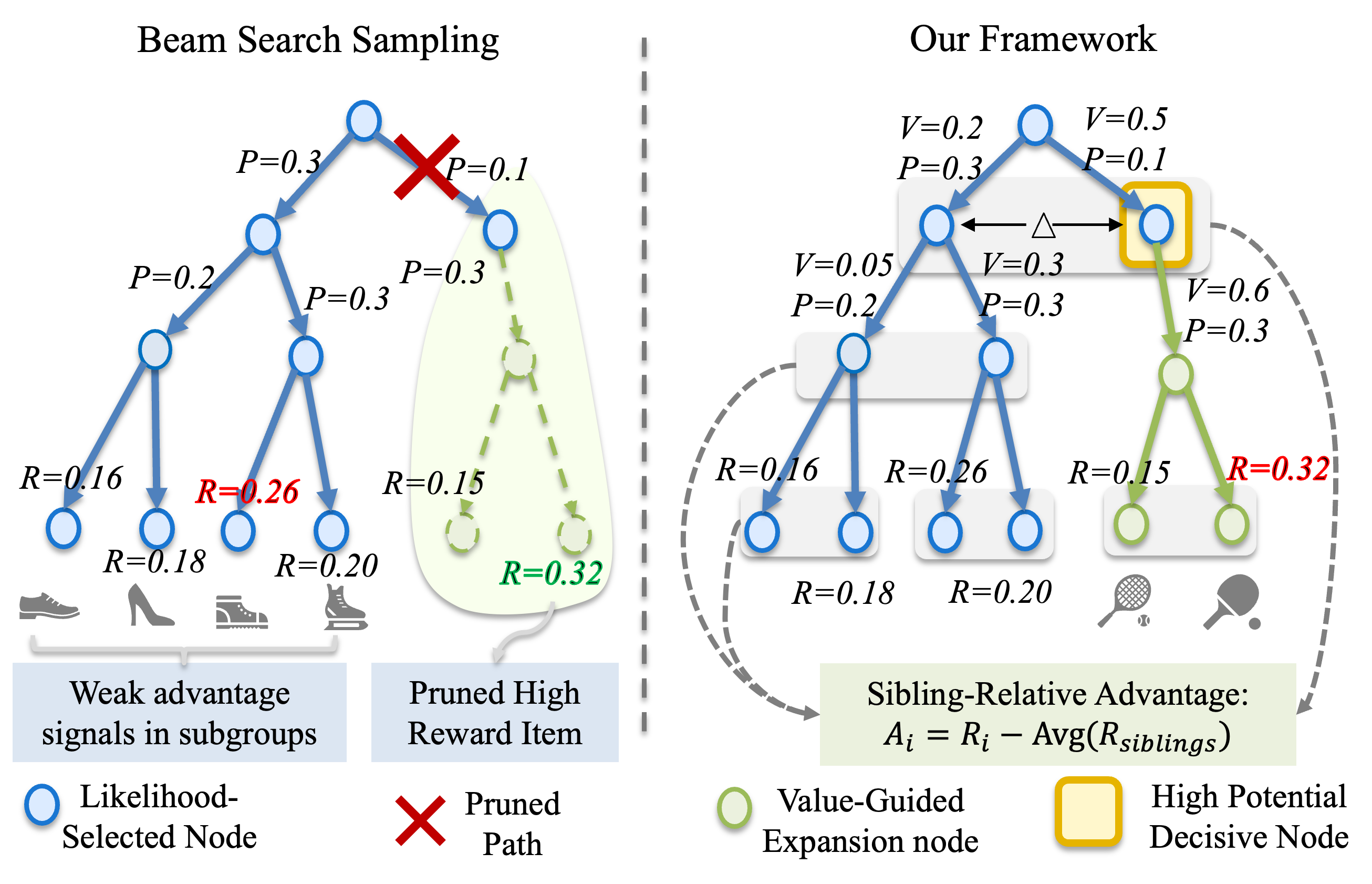}
    \caption{Comparison of Beams Search and V-STAR.  Left: probability-based pruning removes high-reward items and produces homogeneous candidates. Right: V-STAR expands high-value prefixes under a fixed budget and strengthens within-group learning with Sibling-GRPO.}
    \label{fig:introduction}
    \vspace{-0.1em}
\end{figure}

This misalignment creates a tension between the greedy nature of generation and the long-term goal of exploration, manifesting in two critical structural failures: (1) \textbf{insufficient exploration of high-reward items:} Many high-reward items begin with lower-probability tokens because they have limited historical interactions. Standard beam search prunes branches based on likelihood and can irreversibly discard such paths early in generation. Consequently, the model becomes blind to high-reward candidates simply because their prefixes are not immediately likely. (2) \textbf{advantage compression within the generated group:} Guided by likelihood, decoding tends to over-expand multiple branches that share high-probability prefixes, producing a candidate set dominated by siblings, i.e., near-duplicate items with similar rewards. However, effective RL relies on reward contrast: policy gradients are driven by differences between good and bad actions. In such sibling-heavy groups, rewards become highly correlated and concentrate in a narrow band, while a few outliers can set the overall reward scale. As a result, GRPO's group-wise normalization is dominated by the global reward range and shrinks the relative advantage differences among siblings, weakening the learning signal at the decisions that actually matter.

Previous methods attempt to mitigate these issues, but each incurs a hard trade-off. One line of work increases the sampling budget~\cite{goyal2018continuous, wiseman2016sequence, yang2026bear}, yet indiscriminate expansion largely spends compute on redundant, high-probability candidates. Another line relies on heuristic sampling (e.g., temperature scaling or nucleus sampling) to increase sample diversity~\cite{holtzman2019curious, ouyang2022training, touvron2023llama}, but the added randomness is difficult to control, often hurting relevance and making high-reward outcomes less reliable. More structured approaches, such as Monte Carlo Tree Search or tree-of-thought style search~\cite{silver2017mastering, yao2023tree, li2022competition}, can improve exploration in LLM inference, but their computational cost is prohibitive for broad use in recommendation. A critical gap therefore remains: we lack a mechanism that reliably determines \emph{when} exploration is worth paying for and \emph{where} limited compute should be allocated to maximize reward.

Motivated by this gap, we make a simple observation: under a fixed decoding budget, extra exploration does \emph{not} help equally at every generation step. Many prefixes are clearly low-potential, or their next token is already obvious, so additional compute there is largely wasted. In contrast, exploration matters most at a small number of \emph{decisive} prefixes where several plausible branches compete and which branch to choose strongly affects the final reward (e.g., whether long-tail but high-reward items can be reached). We detect these decisive points using two signals: (1) \textbf{high value}, meaning the prefix is promising in expected reward, and (2) \textbf{high ambiguity given high value}, meaning the next-step choice is uncertain enough that extra search can change the outcome. This perspective turns exploration into a targeted budgeting problem: decide \emph{where} to spend compute so that it most improves reward, which directly motivates a value-guided decoding strategy and a tree-structured credit assignment objective.

Building on these insights, we propose \textbf{V-STAR} (\underline{V}alue-guided \underline{S}ampling and \underline{T}ree-structured \underline{A}dvantage \underline{R}einforcement), a self-evolving sampling-and-learning framework where better sampling provides clearer training signals, and the resulting updates further improve sampling. V-STAR comprises two key components: \textbf{Value-Guided Efficient Decoding (VED)} for candidate sampling and \textbf{Sibling-GRPO} for tree-structured policy optimization. \textbf{VED} tackles \textit{insufficient exploration} by performing {budgeted} value-guided expansion on the SID prefix tree. It initializes a shallow prefix tree with likelihood-guided beam search to cheaply reveal where hypotheses concentrate and where promising branches are pruned early. It then combines a lightweight value estimator with an uncertainty signal (e.g., policy entropy) to select a small set of high-potential prefixes and spends additional compute only at these divergence points, enabling local expansion and backup under strict depth and branching constraints. This targeted allocation corrects premature pruning and improves the reachability of long-tail, high-reward items without incurring exhaustive tree search. 
\textbf{Sibling-GRPO} addresses \textit{advantage compression} in prefix-coupled generation by introducing a structured group-relative objective over genealogically related candidates. Rather than applying a single global normalization over the entire candidate set, it forms sibling groups under each parent prefix and performs within-group relative advantage learning. This concentrates gradients on the decisive branching actions where candidates diverge yet remain highly correlated, mitigating advantage compression and recovering informative learning signals inside homogenized candidate clusters. Together, VED improves candidate quality and diversity, while Sibling-GRPO turns these gains into more stable and informative updates. 

In summary, our contributions are as follows: \begin{enumerate} \item We formalize probability--reward misalignment in generative recommendation and analyze its impact on sample decoding and RL fine-tuning. Building on this analysis, we propose V-STAR to address the issue via value-guided decoding and tree-structured credit assignment. \item We develop Value-Guided Efficient Decoding (VED) for budgeted candidate construction by allocating decoding compute to high-value and high-ambiguity nodes. \item We introduce Sibling-GRPO, a tree-structured objective that uses sibling groups to mitigate advantage compression under correlated candidates. \item Experiments on both offline datasets and online settings show consistent gains over strong generative baselines. \end{enumerate}

\section{Related Work}

\subsection{Generative Recommender}
The paradigm of recommender systems is shifting from discriminative matching to generative modeling. Early works linearized user histories and predicted the next item ID directly, but faced challenges due to the extremely large item-ID vocabulary~\cite{covington2016deep, sun2019bert4rec}. To mitigate this issue, recent approaches such as P5~\cite{geng2022recommendation} and TIGER~\cite{rajput2023recommender} introduce SID, which represent each item as a short sequence of hierarchical discrete tokens constructed via residual quantization (RQ-VAE)~\cite{zeghidour2021soundstream} or collaborative indexing. This formulation bridges recommendation and language modeling, enabling Transformer architectures to generate items token by token.
Most existing SID-based recommenders still rely on likelihood-driven decoding (e.g., beam search) at training time. Recent analyses~\cite{li2023generative} show that beam search exhibits a rich-get-richer bias and can prune low-probability branches early in generation. In recommendation, high-reward items (e.g., niche discoveries) may begin with low-probability prefix tokens, so early prefix pruning can make a substantial portion of the item space effectively unreachable. This is a structural limitation that is not resolved by model scaling alone.

% \subsection{Generative Recommender Systems}
% Recommender systems are increasingly shifting from discriminative matching pipelines to generative sequence modeling.
% Early sequence-based recommenders model user behavior as an item-ID sequence and learn to predict future items over a large vocabulary, which poses challenges for both modeling and inference at scale~\cite{covington2016deep,sun2019bert4rec}.
% To mitigate the item-vocabulary bottleneck, recent generative recommenders represent each item using structured discrete tokens, often referred to as Semantic IDs, constructed via hierarchical quantization (e.g., RQ-VAE)~\cite{zeghidour2021soundstream} or collaborative indexing, enabling token-by-token generation with Transformer architectures~\cite{geng2022recommendation,rajput2023recommender}.
% In practice, these methods typically rely on likelihood-dominated decoding such as beam search to generate candidate items under strict latency budgets.
% Recent analyses~\cite{li2023generative} show that beam search exhibits a rich-get-richer behavior, aggressively pruning low-probability branches early in the generation process.
% For recommendation, high-utility items (e.g., niche discoveries) can correspond to low-probability early tokens, so early prefix pruning may prevent these items from being sampled and thus limit the effective coverage of the item space during both inference and self-generated training.

\vspace{-1em}
\subsection{Decoding Strategies in Generation}
Decoding (sampling) strategies control the trade-off between exploration and exploitation. Heuristic sampling methods (e.g., Top-K and nucleus sampling~\cite{holtzman2019curious}) introduce stochasticity to improve diversity, but they provide no explicit mechanism to guarantee reward and can lead to irrelevant recommendations. In contrast, tree-search algorithms such as Monte Carlo Tree Search (MCTS) and Tree of Thoughts (ToT)~\cite{yao2023tree} enable lookahead and backtracking and have shown strong performance on multi-step reasoning tasks. However, these methods are often impractical for industrial recommendation because strict latency budgets make per-request rollouts prohibitively expensive.
Some works distill search behavior into a policy~\cite{wang2023learning}, but they typically treat search as a black-box optimizer. Our work instead proposes a budget-aware sparse search that dynamically allocates a finite compute budget to a small set of pivotal nodes, capturing the benefits of lookahead without the overhead of full tree expansion.

\subsection{RL for Generative Alignment}
RL is widely used to align generative models with non-differentiable objectives (e.g., CTR and diversity). Common approaches include REINFORCE~\cite{williams1992simple}, PPO~\cite{schulman2017proximal}, GRPO~\cite{shao2024deepseekmath}, and recent direct preference-optimization methods such as DPO~\cite{rafailov2024direct}. In recommendation, RL has been applied to optimize slate generation (e.g., SlateQ~\cite{ie2019reinforcement}) and to correct exposure bias~\cite{chen2019top}. 
However, existing generative RL methods often implicitly treat candidate samples as independent and identically distributed (i.i.d.)~\cite{deng2025onerec, wei2025oneloc}. In autoregressive generation, candidates frequently share long prefixes and therefore form sibling groups with highly correlated features and rewards. Standard policy-gradient updates can be ineffective in this setting: subtracting a baseline from a set of nearly identical rewards yields a weak and noisy advantage signal, which can lead to advantage collapse~\cite{ramesh2023policy}. Our proposed Sibling-GRPO addresses this issue by explicitly modeling relative advantages within sibling groups, which is related in spirit to listwise ranking objectives~\cite{cao2007learning} but tailored to the hierarchical structure of the generation tree.

% =========================
% Preliminary (LaTeX Version)
% =========================

\section{Preliminaries}
\label{sec:prelim}
\subsection{Problem Formulation}
\noindent\textbf{Generative Recommendation with Semantic IDs.}\;
We formulate recommendation as a conditional sequence generation task~\cite{sutskever2014sequence, geng2022recommendation}. Under the SID paradigm~\cite{rajput2023recommender}, each item is represented by a unique, fixed-length sequence of discrete tokens
$y=(y_1,\ldots,y_L)\in\mathcal{V}^L$,
where $\mathcal{V}$ is a finite token vocabulary and $L$ is the (fixed) SID length. Given a user context $x$ (e.g., interaction history and profile features), an autoregressive policy $\pi_\theta$ models the conditional probability of an item sequence via the probability chain rule:
$
\pi_\theta(y \mid x)
=
\prod_{l=1}^{L}\pi_\theta\!\left(y_{l} \mid x, y_{\leq l-1}\right),
\label{eq:policy_factorization}
$
where $y_{\leq l-1}=(y_1,\ldots,y_{l-1})$ denotes the already generated prefix before step $l$, with $y_{\leq 0}=\varnothing$ (empty prefix, i.e., generation starts from the user context $x$ alone).

\noindent\textbf{Decoding Operator.}\;
We abstract decoding process as an operator $\mathcal{D}$ that maps policy $\pi_\theta$ and context $x$ to a candidate set $\mathcal{C}(x)$:
\begin{equation}
\mathcal{C}(x)=\mathcal{D}(\pi_\theta,x),\qquad \mathcal{C}(x)\subset \mathcal{V}^L.
\label{eq:decoding_operator}
\end{equation}
Ideally, $\mathcal{D}$ should return candidates that maximize the ground-truth reward $R(x,y)$. In practice, standard decoding strategies (e.g., beam search~\cite{yang2026bear}) approximate this objective by maximizing model probability $\pi_\theta(y\mid x)$ through heuristic prefix pruning.

\noindent\textbf{Policy Refinement via GRPO.}\;
To align the pre-trained policy with non-differentiable ranking objectives, we apply GRPO~\cite{shao2024deepseekmath} to the decoded set $\mathcal{C}(x)$ as the empirical reference set for reward comparison. The standard group-normalized advantage is used:
\begin{equation}
A(x,y)
=
\frac{R(x,y)-\mu_R(x)}{\sigma_R(x)+\epsilon},
\label{eq:advantage}
\end{equation}
where $\mu_R(x)$ and $\sigma_R(x)$ are the mean and standard deviation of rewards over $\mathcal{C}(x)$ and $\epsilon>0$ ensures numerical stability. In addition to standard GRPO, our method also applies sibling GRPO to further enhance learning signals, which will be introduced in Sec.~\ref{sec:sibling_grpo}.

\subsection{Probability-Driven Decoding Bias}
\label{sec:prob_driven_decoding_bias}

\noindent\textbf{Insufficient exploration on high-reward items.}\;
Autoregressive decoding process can be viewed as traversing a prefix tree~\cite{weng2025traversal}: SIDs sharing a common prefix follow the same branch until they diverge. Under a fixed budget, pruning-based methods~\cite{ouyang2022training, cobbe2021training, touvron2023llama} (e.g., beam search or top-K sampling) keep only a small set of active prefixes and discard the rest by local likelihood. Pruning is irreversible in this case: once a prefix is removed, all items under that branch will never be evaluated. This is problematic because likelihood and reward can be misaligned: there may exist $y_a,y_b$ with $\pi_\theta(y_a\mid x)>\pi_\theta(y_b\mid x)$ but $R(x,y_a)<R(x,y_b)$. As a result, likelihood-only pruning may eliminate low-probability prefixes that lead to high-reward items, and those branches are rarely revisited. %This motivates value-guided budget allocation, which invests compute in prefixes with high predicted return rather than pruning solely by likelihood.

% \begin{figure}[t]
%     \centering
%     \includegraphics[width=\linewidth]{figure/adv_colla.png}
%     \caption{The phenomenon of Advantage Compression}
%     \label{fig:adv_colla}
%     \vspace{-0.8em}
% \end{figure}

\noindent\textbf{Advantage compression within the generated group.}\;
Given a candidate set $\mathcal{C}(x)$, we define the within-group reward range as:
\begin{equation}
\Delta_R(x)
\triangleq
\max_{y\in \mathcal{C}(x)} R(x,y)
-
\min_{y\in \mathcal{C}(x)} R(x,y).
\label{eq:reward_range}
\end{equation}
Since $|R(x,y)-\mu_R(x)|\le \Delta_R(x)$ for all $y\in\mathcal{C}(x)$, the within-group advantage magnitude is bounded by:
\begin{equation}
|A(x,y)|
=
\left|
\frac{R(x,y)-\mu_R(x)}{\sigma_R(x)+\epsilon}
\right|
\le
\frac{\Delta_R(x)}{\sigma_R(x)+\epsilon}
\le
\frac{\Delta_R(x)}{\epsilon}.
\label{eq:adv_bound_range}
\end{equation}
Eq.~\ref{eq:adv_bound_range} implies that $A(x,y)\in[-\frac{\Delta_R(x)}{\epsilon},\frac{\Delta_R(x)}{\epsilon}]$.
By Popoviciu's inequality on variances, $\mathrm{Var}_{y\in\mathcal{C}(x)}[A(x,y)]\le (\frac{\Delta_R(x)}{\epsilon})^2$, so the standard deviation
$
\sigma_A(x)
\triangleq
\sqrt{\mathrm{Var}_{y\in\mathcal{C}(x)}[A(x,y)]}
\le
\frac{\Delta_R(x)}{\epsilon}.
\label{eq:sigmaA_popoviciu}
$
Therefore, when candidate collapse in $\mathcal{C}(x)$ yields small $\Delta_R(x)$, it forces $\sigma_A(x)$ to be small, making advantages nearly indistinguishable.
We refer to this degeneration of learning signal as \emph{advantage compression}.

\noindent\textbf{Implication.}\;
Together, insufficient exploration and advantage compression show that decoding is not a mere inference-time heuristic~\cite{shi2024thorough}: it controls both \emph{reachability} (which high-reward items are ever surfaced) and \emph{learnability} (whether candidates provide enough reward contrast for RL). We therefore need a decoder that avoids irreversible pruning of low-probability yet high-reward branches and yields candidate groups with sufficient reward dispersion, motivating our dynamic decoding framework.

\begin{figure*}[t]
    \centering
    \includegraphics[width=\linewidth]{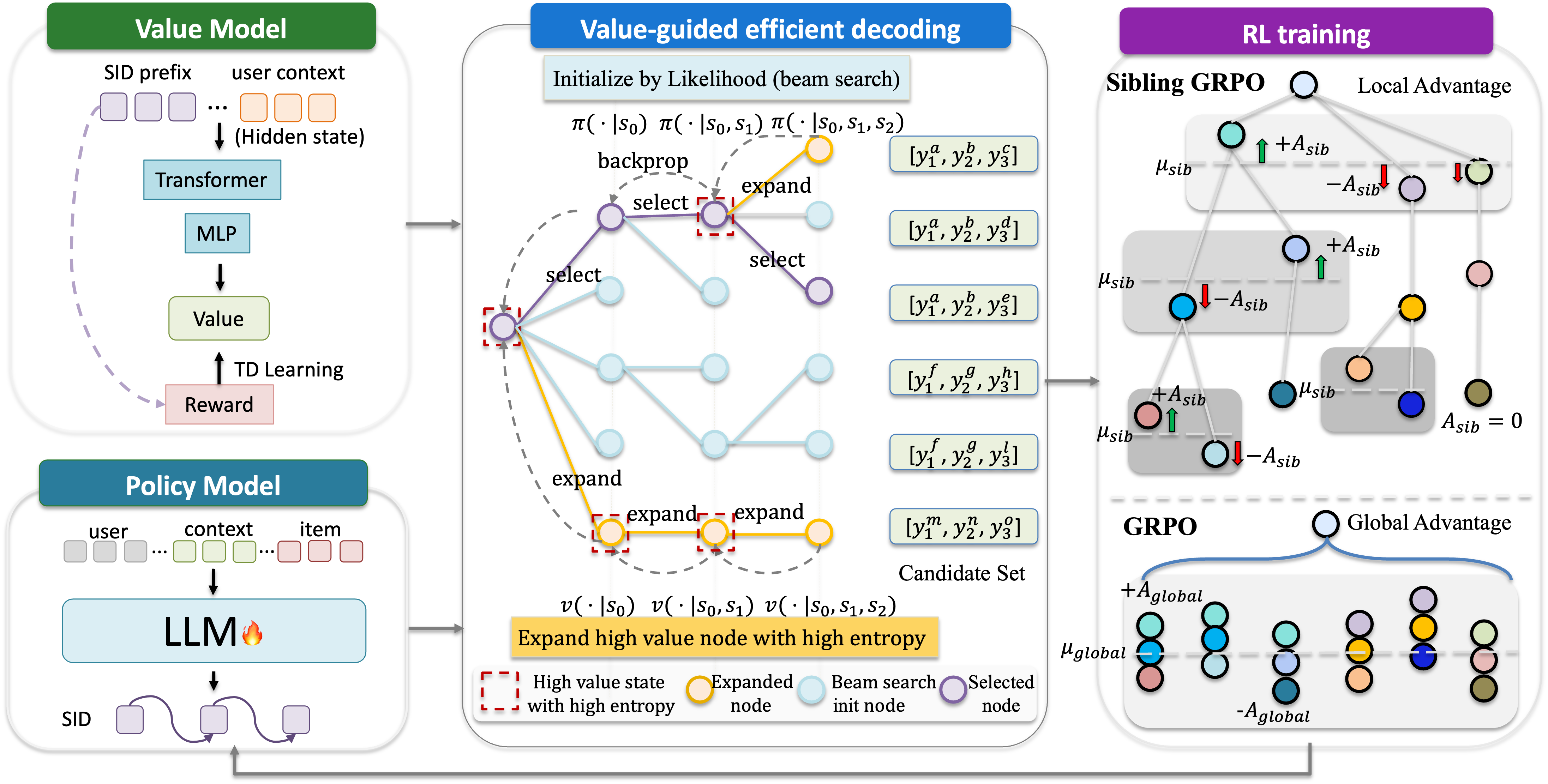}
    \caption{Overview of the proposed self-evolving decoding-and-learning framework V-STAR.}
    \label{fig:backbone}
    \vspace{-0.8em}
\end{figure*}

\section{Methodology}
\label{sec:method}
Our V-STAR (Fig.~\ref{fig:backbone}) is a unified decoding-and-learning framework consisting \emph{Value-Guided Efficient Decoding} (VED) for candidate construction and \emph{Sibling-GRPO} for policy optimization. VED addresses \emph{reachability} by allocating compute to decisive prefixes under a fixed budget, while Sibling-GRPO addresses \emph{learnability} by restoring within-group learning signal under prefix coupling. 

\vspace{-1em}
\subsection{Semantic-aware Value Model Learning}
\label{sec:value_learning}
VED relies on lookahead value estimation to mitigate probability-driven bias on the SID prefix tree. To this end, we first learn a prefix value function $V_\phi$ that maps each prefix to its expected downstream return~\cite{schulman2017proximal}. At generation step $\ell\in\{1,\ldots,L\}$, we define the decoding state $s_\ell$ as
$
s_\ell \triangleq (x, y_{\leq \ell}),
$
where $x$ is the user context and $y_{\leq \ell}=(y_1,\ldots,y_{\ell})$ is the current SID prefix. The value function estimates the discounted return of continuing generation from $s_\ell$:
\begin{equation}
V_\phi(s_\ell)
\triangleq
\mathbb{E}\!\left[\sum_{t=\ell}^{L}\gamma^{\,t-\ell}\, r_t \ \middle|\ s_\ell\right],
\label{eq:value_return}
\end{equation}
where $\gamma\in(0,1]$ is the discount factor. We parameterize $V_\phi$ as a lightweight value head on top of the policy backbone: a single shallow Transformer block (operating on the prefix hidden states) followed by an MLP regressor, enabling frequent value queries with negligible overhead.

%ground-truth SID $y^\star=(y_1^\star,\ldots,y_L^\star)$ and pre-computed 

\noindent\textbf{Semantic-aware dense supervision.}\;
Learning $V_\phi$ from sparse terminal returns (non-zero only when the generated item exactly matches the ground truth) provides weak, delayed signals for intermediate prefixes. We therefore build dense semantic returns from pre-computed item embeddings, obtained by encoding each item’s textual description with a frozen text encoder. Given context $x$, let $\mathcal{C}(x)=\mathcal{D}(\pi_\theta,x)$ be the sampled candidate set. For prefix $y_{\leq \ell}$, we define the \emph{sampled prefix bucket} to be the subset of items in $\mathcal{C}(x)$ that shares this prefix:
\begin{equation}
\mathcal{C}(x; y_{\leq \ell})
\triangleq
\left\{\, y^\prime \in \mathcal{C}(x) \;|\; y^\prime_{\leq \ell} = y_{\leq \ell} \,\right\},
\label{eq:sampled_prefix_bucket}
\end{equation}
where $y^\prime_{\leq \ell}$ denotes the SID prefix of the candidate item $y^\prime$.
We then define a context-conditional \emph{prefix embedding} by averaging embeddings over this bucket:
\begin{equation}
\bar{\mathbf{e}}(x, y_{\leq \ell})
\triangleq
\frac{1}{|\mathcal{C}(x; y_{\leq \ell})|}
\sum_{i \in \mathcal{C}(x; y_{\leq \ell})} \mathbf{e}(i),
\label{eq:avg_prefix_emb}
\end{equation}
where $\mathbf{e}(i)$ is the embedding of candidate item $i$. The step-wise dense return $r_\ell$ for prefix $y_{\leq \ell}$ is then defined as:
\begin{equation}
r_\ell \;=\;
\begin{cases}
w_\ell, 
& \text{if } y_{\leq \ell}=y^\star_{\leq \ell}, \\[4pt]
-\,w_\ell\!\left(1-\cos\!\big(\bar{\mathbf{e}}(x, y_{\leq \ell}), \mathbf{e}(y^\star)\big)\right),
& \text{otherwise},
\end{cases}
\label{eq:step_reward}
\end{equation}
where $w_\ell>0$ is a monotonically increasing weight with respect to $\ell$, $\cos$ denotes the cosine similarity between the embeddings, and $\mathbf{e}(y^\star)$ represents the embedding of the ground-truth item $y^\star$ for context $x$. Compared to sparse exact-match supervision, the proposed semantic-aware dense supervision provide informative feedback on mismatched prefixes by leveraging their semantic proximity to the ground-truth item, thereby improving value estimation for intermediate decoding states.

% \paragraph{Temporal-difference learning.}
% Given the stepwise signal in~\eqref{eq:step_reward}, we train the critic by one-step temporal-difference learning, which enforces Bellman consistency across successive prefixes. With a target network $\phi'$, the TD target and regression loss are
% \begin{equation}
% \tilde y_\ell \;=\; r_\ell \;+\; \gamma\, V_{\phi'}(s_{\ell+1}),
% \qquad
% \mathcal{L}_{V}(\phi)
% \;=\;
% \mathbb{E}\Big[\big(V_\phi(s_\ell)-\tilde y_\ell\big)^2\Big],
% \label{eq:td_learning}
% \end{equation}
% where $s_{\ell+1}=(x,y_{<\ell+1})$. The resulting $V_\phi$ provides a cheap, reward-aligned lookahead over prefixes, which is later used to prioritize budgeted expansion toward high-upside decision points in Sec.~\ref{sec:VED}.

\noindent\textbf{Temporal-difference learning.}\;
Given the stepwise signal in~\eqref{eq:step_reward}, we train the value function by one-step temporal-difference learning, enforcing Bellman-style consistency across successive prefixes. Concretely, the TD target and regression loss are:
\begin{equation}
\tilde y_\ell \;=\;
\begin{cases}
r_\ell \;+\; \gamma\, V_{\phi}(s_{\ell+1}), & \ell < L, \\[4pt]
r_L, & \ell = L,
\end{cases}
\qquad
\mathcal{L}_{V}(\phi)
\;=\;
\mathbb{E}\Big[\big(V_\phi(s_\ell)-\tilde y_\ell\big)^2\Big],
\label{eq:td_learning}
\end{equation}
where $s_{\ell+1}=(x,y_{\leq \ell+1})$. The resulting $V_\phi$ provides an efficient, reward-aligned lookahead over prefixes, which is later used to guide budgeted expansion toward decisive prefix nodes in Sec.~\ref{sec:VED}.

\vspace{-1em}
\subsection{Value-Guided Efficient Decoding}
\label{sec:VED}
With the value model as guidance, we cast semantic-ID decoding as constructing a candidate set under a strict compute budget on the SID prefix tree. Given context $x$, the decoder returns $\mathcal{C}(x)\subset\mathcal{V}^L$ that (i) contains high-reward items and (ii) enables informative within-group comparisons for groupwise RL.

\noindent\textbf{A budgeted objective for candidate sets.}\;
Let $\mathrm{Cost}(\mathcal{C}(x))$ denote the compute required to construct $\mathcal{C}(x)$, measured by the number of backbone forward tokens incurred during decoding.
We formulate candidate construction as a constrained set optimization problem:
\begin{equation}
\begin{aligned}
\max_{\mathcal{C}(x)\subseteq \mathcal{V}^L}\quad
& \underbrace{\mathbb{E}_{y\in \mathcal{C}(x)}[R(x,y)]}_{\text{reward}}
\;+\;
\lambda\,\underbrace{\mathrm{Contrast}(\mathcal{C}(x))}_{\text{dispersion}} \\
\text{s.t.}\quad
& \mathrm{Cost}(\mathcal{C}(x)) \le B.
\end{aligned}
\label{eq:set_objective}
\end{equation}
where $\mathrm{Contrast}(\cdot)$ is a set-level regularizer that encourages within-group discriminability, e.g., by increasing reward dispersion and mitigating prefix coupling (Sec.~\ref{sec:prob_driven_decoding_bias}). The budget $B$ captures the computational constraint in recommendation.

\noindent\textbf{Value-based scoring with entropy regularization.}\;
Directly optimizing Eq.~\eqref{eq:set_objective} is intractable due to the exponential SID space. VED therefore allocates the decoding budget to a small set of \emph{decisive prefixes}. The key question is how to rank prefix states $s=(x,y_{\le \ell})$ for further expansion. To maximize the expected improvement per unit cost, prefix expansion should balance \emph{exploitation} and \emph{exploration}. We therefore define a prefix priority score:
\begin{equation}
G(s) \ \triangleq\
\begin{cases}
V_\phi(s) \;+\; \lambda\,\mathcal{H}_\theta(s), & \ell < L, \\[4pt]
V_\phi(s), & \ell = L,
\end{cases}
\label{eq:acq}
\end{equation}
where $V_\phi(s)$ is the predicted downstream return from prefix $s$, $\lambda$ is the regularization weight, and $\mathcal{H}_\theta(s)$ is the entropy term that measures how uncertain the policy is about the next token:
\begin{equation}
\mathcal{H}_\theta(s) \ \triangleq\ -\sum_{y_{\ell+1}\in\mathcal{V}} \pi_\theta(y_{\ell+1}\mid x,y_{\leq \ell})\log \pi_\theta(y_{\ell+1}\mid x,y_{\leq \ell}).
\label{eq:entropy}
\end{equation}
For terminal prefixes ($\ell=L$), the priority reduces to $V\phi(s)$.

\noindent\textbf{Score-based efficient decoding.}\;
\label{sec:VED_solver}
With the above value-based prefix score, the decoding process performs budgeted search and expansion on a search tree $\mathcal{T}$ over prefix states $s=(x,y_{\leq \ell})$.  The procedure consists of four stages: \textcircled{1} initialization, \textcircled{2} selection and traversal, \textcircled{3} gated expansion, and \textcircled{4} backpropagation.

\textit{\textcircled{1} Initialization.}
We initialize $\mathcal{T}$ using prefixes obtained by a low-cost probability-guided beam search under $\pi_\theta$. The beam width in this warm start is set smaller than the final candidate set size to limit overhead. The resulting prefix tree reveals which prefixes the policy prefers and which branches are pruned early. We evaluate each node in the initial tree once to populate $\{V_\phi(s), \mathcal{H}_\theta(s), G(s)\}$ and record whether the node admits unexpanded valid children.

\textit{\textcircled{2} Selection and traversal.}
After initialization, we repeatedly select nodes and traverse a root-to-leaf path. Let $N(s)$ be the visit count of node $s$ and $N_{\mathrm{root}}$ the number of traversals so far, we select nodes with a UCB-style score~\cite{kocsis2006bandit}:
\begin{equation}
U(s)
\triangleq
G(s)
\;+\;
\beta \cdot \sqrt{\frac{\ln\!\big(N_{\mathrm{root}}+1\big)}{N(s)+1}},
\label{eq:VED_select}
\end{equation}
where $\beta$ controls exploration. Starting from the root, we recursively choose the child with the largest $U(s)$ until reaching a leaf or terminal node.

\textit{\textcircled{3} Gated expansion.}
Along the selected path, we expand only prefixes that are decisive relative to other prefixes at the same depth. Let $\mathcal{T}_\ell$ denote the set of nodes in the current tree at depth $\ell$, and define the depth-wise average priority
\begin{equation}
\bar{G}_\ell
\triangleq
\frac{1}{|\mathcal{T}_\ell|}
\sum_{u\in\mathcal{T}_\ell} G(u).
\label{eq:depth_avg}
\end{equation}
For any visited node $s=(x,y_{\leq \ell})$ on the path that is not fully expanded, we trigger one-step expansion if and only if
$
G(s)\ \ge\ \bar{G}_\ell.
\label{eq:VED_gate}
$
%restrict the action space to the top-$q$ valid next tokens under $\pi_\theta(\cdot\mid s)$ among yet-unexpanded children. We then 
When triggered, we add one new child to $s$ by sampling a token from its yet-unexpanded valid children according to their normalized policy probabilities.

\textit{\textcircled{4} Backpropagation.}
For each newly added child $s_{\mathrm{new}}$, we evaluate $(V_\phi,\mathcal{H}_\theta,G)$ and update statistics along its ancestor chain, including visit count $N(\cdot)$ and depth-wise average priority scores.

The above stages \textcircled{2}--\textcircled{4} are repeated until $\mathrm{Cost}(\mathcal{T}) > B$, where $\mathrm{Cost}(\cdot)$ counts the number of backbone forward tokens. After termination, we extract a candidate set $\mathcal{C}(x)\subset\mathcal{V}^L$ from the search tree $\mathcal{T}$ in a value-aware manner. In practice, we sample top-valued SIDs from depth-$L$ leaf nodes. In rare cases, if insufficient leaf nodes exist, we fill the remainder by completing high-value prefixes in the tree using cached $\pi_\theta$, without additional backbone forwards. The resultant candidate set $\mathcal{C}$ improves reachability by exploring low-probability but high-value prefixes.

\newcommand{\cmark}{\scalebox{0.85}{\ding{51}}}

\begin{table*}[t]
\centering
\small
\caption{Performance comparison on Industrial and Office. Gen. means Generative model. Best results are \textbf{bolded}.}
\setlength{\tabcolsep}{2.7pt}
\begin{tabular}{l c c c c c c c c c c c c c c c}
\toprule
\cmidrule(lr){5-10}\cmidrule(lr){11-16}
\multirow{2}{*}{{Model}} & \multirow{2}{*}{{Gen.}} & \multirow{2}{*}{{SID}} & \multirow{2}{*}{{RL}}& \multicolumn{6}{c}{Industrial} & \multicolumn{6}{c}{Office} \\
\cmidrule(lr){5-10}\cmidrule(lr){11-16}
& & & 
& HR@3 & HR@5 & HR@10 & NDCG@3 & NDCG@5 & NDCG@10
& HR@3 & HR@5 & HR@10 & NDCG@3 & NDCG@5 & NDCG@10 \\
\midrule

GRU4Rec  & $-$ & $-$ & $-$
& 0.0638 & 0.0774 & 0.0999 & 0.0542 & 0.0598 & 0.0669
& 0.0629 & 0.0789 & 0.1019 & 0.0528 & 0.0595 & 0.0669 \\
Caser    & $-$ & $-$ & $-$
& 0.0618 & 0.0717 & 0.0942 & 0.0514 & 0.0555 & 0.0628
& 0.0748 & 0.0865 & 0.1093 & 0.0615 & 0.0664 & 0.0737 \\
SASRec   & $-$ & $-$ & $-$
& 0.0790 & 0.0909 & 0.1088 & 0.0700 & 0.0748 & 0.0806
& 0.0861 & 0.0949 & 0.1120 & 0.0769 & 0.0805 & 0.0858 \\
\midrule

HSTU     & \cmark & $-$ & $-$
& 0.0927 & 0.1037 & 0.1163 & 0.0885 & 0.0918 & 0.0958
& 0.1134 & 0.1252 & 0.1400 & 0.1031 & 0.1079 & 0.1126 \\
BIGRec   & \cmark & $-$ & $-$
& 0.0931 & 0.1092 & 0.1370 & 0.0841 & 0.0907 & 0.0997
& 0.1069 & 0.1204 & 0.1434 & 0.0961 & 0.1017 & 0.1091 \\
\midrule

TIGER    & \cmark & \cmark & $-$
& 0.0852 & 0.1010 & 0.1321 & 0.0742 & 0.0807 & 0.0908
& 0.0986 & 0.1163 & 0.1408 & 0.0852 & 0.0960 & 0.1002 \\
LCRec    & \cmark & \cmark & $-$
& 0.0915 & 0.1057 & 0.1332 & 0.0805 & 0.0862 & 0.0952
& 0.0921 & 0.1048 & 0.1237 & 0.0807 & 0.0859 & 0.0920 \\
D3       & \cmark & \cmark & $-$
& 0.1024 & 0.1213 & 0.1500 & 0.0991 & 0.0989 & 0.1082
& 0.1204 & 0.1406 & 0.1634 & 0.1055 & 0.1139 & 0.1213 \\
\midrule

S-DPO    & \cmark & \cmark & \cmark
& 0.1032 & 0.1238 & 0.1524 & 0.0906 & 0.0991 & 0.1082
& 0.1169 & 0.1356 & 0.1587 & 0.1033 & 0.1110 & 0.1255 \\
MiniOneRec
         & \cmark & \cmark & \cmark
& 0.1143 & 0.1321 & 0.1586 & 0.1011 & 0.1084 & 0.1167
& 0.1217 & 0.1420 & 0.1634 & 0.1088 & 0.1172 & 0.1242 \\
\midrule

\textbf{Ours-Train}
         & \cmark & \cmark & \cmark
& \textbf{0.1189} & \textbf{0.1361} & \textbf{0.1641} & \textbf{0.1057} & \textbf{0.1128} & \textbf{0.1217}
& \textbf{0.1344} & \textbf{0.1500} & \textbf{0.1746} & \textbf{0.1196} & \textbf{0.1260} & \textbf{0.1340} \\

% \textbf{Ours}
%          & \cmark & \cmark & \cmark
% & \textbf{0.1204} & \textbf{0.1366} & \textbf{0.1650} & \textbf{0.1066} & \textbf{0.1141} & \textbf{0.1233}
% & \textbf{0.1353} & \textbf{0.1518} & \textbf{0.1760} & \textbf{0.1213} & \textbf{0.1257} & \textbf{0.1352} \\

\bottomrule
\end{tabular}
\vspace{-0.8em}
\label{tab:perf_compare}
\end{table*}

\subsection{Sibling-GRPO}
\label{sec:sibling_grpo}
While VED improves reachability and increases candidate reward dispersion, hierarchical SIDs still induce common-prefix sibling node groups in $\mathcal{C}(x)$. Sibling-GRPO exploits these groups to recover strong learning signals exactly at the decisive branching actions.

\noindent\textbf{Sibling groups and sibling nodes.}\;
An SID $y=(y_1,\ldots,y_L)$ corresponds to a root-to-leaf path on a depth-$L$ prefix tree. Fix a depth $\ell\in\{1,\ldots,L\}$ and let $h\in\mathcal{V}^{\ell-1}$ denote a parent prefix at depth $\ell-1$. We first define the \emph{sibling group} under $h$ as the sampled items that share this parent prefix:
\begin{equation}
\mathcal{G}(h)
\ \triangleq\
\left\{\, y\in \mathcal{C}(x)\ \middle|\ y_{\leq \ell-1}=h \,\right\}.
\label{eq:cand_group}
\end{equation}
Next, we define the \emph{sibling node set} under $h$ as the set of child nodes at depth $\ell$ that extend $h$ by one token:
\begin{equation}
\mathcal{S}(h)
\ \triangleq\
\left\{\, v\ |\ v\in\mathcal{V},\ \exists\, y\in\mathcal{G}(h)\ \text{s.t.}\ y_\ell=v \,\right\}.
\label{eq:sibling_nodes}
\end{equation}
\noindent\textbf{Relative advantages over sibling nodes.}\;
Sibling-GRPO is computed within $\mathcal{S}(h)$ by assigning each child node $v\in\mathcal{S}(h)$ a node-level score defined as the average reward of candidates routed through that child:
\begin{equation}
\begin{aligned}
\bar{R}(x;h,v)
\ &\triangleq\
\frac{1}{|\mathcal{G}(h,v)|}
\sum_{y\in \mathcal{G}(h,v)} R(x,y), \\
\mathcal{G}(h,v)
\ &\triangleq\
\{\, y\in\mathcal{G}(h)\mid y_\ell=v \,\}.
\end{aligned}
\label{eq:node_avg_reward}
\end{equation}
This construction yields localized, sibling-level comparisons that focus learning on the branching decision at depth $\ell$. Let $\mu_h(x)$ and $\sigma_h(x)$ denote the mean and standard deviation of $\{\bar{R}(x;h,v)\}_{v\in\mathcal{S}(h)}$, the sibling-relative node advantage for a child $v \in\mathcal{S}(h)$ is
\begin{equation}
A_{\mathrm{node}}(x;h,v)
=
\frac{\bar{R}(x;h,v)-\mu_h(x)}{\sigma_h(x)+\epsilon}.
\label{eq:node_adv}
\end{equation}
Compared with global normalization over $\mathcal{C}(x)$, this sibling-node advantage prevents within-parent advantages from being washed out by coarse variations across unrelated prefixes, and it concentrates learning signals on the branching action $y_\ell$ that differentiates siblings under the same parent prefix.

\noindent\textbf{Objective.}\;
Sibling-GRPO retains the GRPO update form but applies it to the sibling nodes within each depth. For a parent prefix $h\in\mathcal{V}^{\ell-1}$ at depth $\ell-1$ and child node $v \in\mathcal{S}(h)$. We define the token-level importance ratio as 
$
\rho_\theta(v\!\mid\!x,h)
\ \triangleq\
\frac{\pi_\theta(v\mid x,h)}{\pi_{\theta_{\mathrm{old}}}(v\mid x,h)},
\label{eq:token_ratio}
$
where $\pi_{\theta_{\mathrm{old}}}$ denotes the frozen behavior policy used to sample $\mathcal{C}(x)$. We then maximize the following sibling-node GRPO objective aggregated over depths:
\begin{equation}
\mathcal{J}_{\mathrm{sib}}(\theta)
=
\mathbb{E}_x
\left[
\frac{1}{|\mathcal{C}(x)|}
\sum_{\ell=1}^{L}
\sum_{h}
\sum_{v\in\mathcal{S}(h)}
A_{\mathrm{node}}(x;h,v)\,
\rho_\theta(v\!\mid\!x,h)
\right].
\label{eq:sib_grpo}
\end{equation}
This formulation performs GRPO-style~\cite{shao2024deepseekmath} updates on branching tokens at each depth, with sibling-normalized advantages computed across competing child nodes under the same parent prefix.

\vspace{-1em}
\subsection{Training and Testing}
\label{sec:self_evolving}
\textbf{Training-Time Self-Evolution.}\; Our V-STAR primarily focuses on \emph{training-time} improvement.
During training, VED and Sibling-GRPO form a closed loop: VED uses the value $V_\phi$ together with the entropy signal to selectively refine search and generate structured candidate groups with better reachability and clearer reward contrast. Sibling-GRPO then converts these structured groups into stable, high-quality policy updates by comparing sibling alternatives at each branching level. As training progresses, both $\pi_\theta$ and $V_\phi$ improve, which sharpens the acquisition score in Eq.~\eqref{eq:acq} and further improves subsequent candidate construction—yielding a self-evolving decoding-and-learning loop.

\noindent\textbf{Testing-Time Decoding.}\;
At test time, we use standard beam search~\cite{sutskever2014sequence} decoding for efficiency and compatibility with existing serving pipelines~\cite{deng2025onerec}.
Optionally, decoding with VED can also be applied at inference: the learned $V_\phi$ and the same acquisition rule Eq.~\eqref{eq:acq} guide the decoder to select a small number of high-potential prefixes under a fixed compute budget, improving long-tail coverage and candidate diversity when additional search is allowed.

% \begin{algorithm}[t]
% \caption{V-STAR Training Loop: VED Sampling + Sibling-GRPO Update}
% \label{alg:ved_sibgrpo_short}
% \KwIn{Policy $\pi_\theta$, value $V_\phi$, token budget $B$; beam width $W$}
% \For {each training iteration}{
%   Sample a minibatch of contexts $\{x\}$\;
%   \For{Each $x$}{
%     \tcp{value-guided candidate construction}
%     $\mathcal{T}\leftarrow$ Initialize Tree with Beam Search
%     \While{$\mathrm{Cost}(\mathcal{T})<B$ \textbf{and} $\mathcal{F}(\mathcal{T})\neq\emptyset$}{
      
%       $\mathcal{T}\leftarrow$ Traversal and expand node with Eq.~\ref{eq:VED_gate} \;
%       { update statistics along its ancestor chain}
%     }
%     $\mathcal{C}(x)\leftarrow$ ExtractLeaves$(\mathcal{T})$ a\;
%     Obtain rewards $\{R(x,y)\}_{y\in\mathcal{C}(x)}$\;

%     \tcp{Update $\theta$ using GRPO with Sibling GRPO}
%   }
% }
% \end{algorithm}

\section{Experiments}
% We conduct a series of experiments to validate the effectiveness of our proposed framework. Our evaluation is designed to answer the following key research questions:

% RQ1: Overall Performance. Does our proposed framework (VED + Sibling-GRPO) achieve state-of-the-art performance compared to strong generative and non-generative baselines?

% RQ2: Impact of Decoding Strategy. How do the different value-guided decoding strategies contribute to the final performance compared to a standard beam search?

% RQ3: Qualitative Analysis. Can we qualitatively observe how our value-guided search discovers better recommendations than purely probability-based decoding?

% RQ4: Importance of the Training Objective. How crucial is the Sibling-GRPO objective for stable and effective policy optimization compared to the standard GRPO?

% Preamble:
% \usepackage{amssymb}  % for \checkmark

% put this in preamble (or right before the table)

% \vspace{3pt}

\subsection{Experimental Setup}
\textbf{Datasets and Evaluation.}\;
Experiments are conducted on both offline public dataset and online data. For offline evaluation, we use Amazon Review dataset~\cite{hou2024bridging} with two subsets \textit{Industrial} and \textit{Office Products}. Following the standard next-item prediction protocol, interactions are ordered chronologically per user. The model takes the entire historical sequence as input and predicts the next interacted item. For data preprocessing, we follow the procedure described in ~\cite{kong2025minionerec}. Each item is represented by a 3-level SID derived from the RQ-VAE codebook~\cite{rajput2023recommender}, and all generative decoders operate under a hard constraint that restricts outputs to valid SIDs, ensuring that each generated sequence maps to a unique catalog item. We report standard metrics Hit Rate (HR) and Normalized Discounted Cumulative
Gain (NDCG). HR@K measures whether the ground-truth next item appears in the Top-$K$ list, while NDCG@K further accounts for its ranked position. Unless otherwise specified, we set $K=3, 5, 10$ in the main tables. For online evaluation, we use real traffic data, which will be detailed in Sec.~\ref{subsec:online_performance}.

\noindent\textbf{Implementation Details.}\;
We use Qwen2.5-1.5B~\cite{qwen2.5} as the model backbone. All models are trained in two stages: supervised fine-tuning (SFT) followed by RL with GRPO. SFT is run with data parallelism on 8 GPUs with a global batch size of 1024 and a fixed random seed. RL is trained on 8 GPUs with a per-GPU batch size 64 and gradient accumulation step 2. The learning rate is set to $1\times10^{-5}$, and KL regularization coefficient is $10^{-3}$. During training, {VED} is utilized to construct $16$ candidates per query, with a prefix length $L=3$ and hierarchical weights $w_l = [0.3, 0.5, 1.0]$. The discount factor $\gamma$ in Eq.~\ref{eq:td_learning} is set to $0.99$ and the exploration coefficient $\lambda$ in Eq.~\ref{eq:acq} is set to $0.1$. The beam width for VED prefix-tree initialization is set to 8. To ensure serving efficiency, we adopt standard {beam search} as the default decoder during inference, effectively decoupling the exploration-intensive training from latency-critical deployment.

\vspace{-1.2em}
\subsection{Performance on Offline Dataset}
The proposed method is compared with recent state-of-the-art baselines under the same setting, including traditional embedding-based sequential recommenders (GRU4Rec \cite{GRU4Rec}, Caser \cite{Caser}, and SASRec \cite{SASRec}), generative non-SID baselines (HSTU \cite{HSTU-ICML-2024}, BIGRec \cite{BIGRec-TORS-2025}), SID-based generative recommenders without RL-style optimization (TIGER \cite{rajput2023recommender}, LCRec \cite{LC-Rec-ICDE-2024}, D3 \cite{D3}), and SID-based methods with RL or preference optimization (S-DPO \cite{kong2025sdpo}, MiniOneRec~\cite{kong2025minionerec}). Table~\ref{tab:perf_compare} summarizes results on \textit{Industrial} and \textit{Office}. Our method consistently achieve the highest performance among all methods. Compared with the strongest RL baseline MiniOneRec, our method achieves a 4.0\% relative improvement in HR@3 and a 4.3\% relative improvement in NDCG@10 on \textit{Industrial}. On \textit{Office}, the gain is more pronounced, reaching a 10.4\% relative improvement in HR@3. These consistent gains indicate that value-guided, budgeted exploration can reliably surface higher-reward SID candidates under strict prefix constraints, while tree-structured advantage reinforcement strengthens training by concentrating gradients on the branching decisions that actually differentiate sibling items.

\vspace{-1.2em}
% \subsection{Online Performance}
% \label{subsec:online_performance}
% To validate V-STAR in a real business environment, we conduct online A/B testing for 7 days on 5\% of live request traffic on \textit{WeChat Channels}. We use \texttt{GMV} (Gross Merchandise Volume) as the primary metric, i.e., the total transaction value attributed to advertising recommendations and a core indicator of commercial revenue. We compare against the \textit{BeamSearch}$+$\textit{GRPO} baseline. The online results show that V-STAR achieves a {0.81\%} relative improvement in \texttt{GMV} and a {1.01\%} relative improvement in \texttt{GMV-Normal} (GMV from click/conversion-optimized ad). These gains suggest that value-aware candidate construction, together with Sibling-GRPO, helps V-STAR more reliably surface high-commercial-value items, thereby increasing transaction volume.

\subsection{Online Performance}
\label{subsec:online_performance}
To validate V-STAR in a real business environment, we conduct online A/B testing for 5 days on 5\% of live request traffic on \textit{WeChat Channels}. We use \texttt{GMV} (Gross Merchandise Volume) as the primary metric, i.e., the total transaction value attributed to advertising recommendations and a core indicator of commercial revenue. We compare against the \textit{BeamSearch}$+$\textit{GRPO} baseline. The online results show that V-STAR achieves a {1.23\%} relative improvement in \texttt{GMV} and a {1.87\%} relative improvement in \texttt{GMV-Normal} (GMV from click/conversion-optimized ad). These gains suggest that value-aware candidate construction, together with Sibling-GRPO, helps V-STAR more reliably surface high-commercial-value items, thereby increasing transaction volume.

% \vspace{-1.2em}
% \subsection{Online Performance}
% \label{subsec:online_performance}
% To validate the advantage of V-STAR in actual business environment, experiments are conducted with real request traffic. Following the paradigm in GPR~\cite{Zhang2025GPR}, \texttt{final\_value} is used as the core evaluation metric, which is a comprehensive value score integrating business indicators such as eCPM, pCTR, and pCVR. Performance is compared against the \textit{BeamSearch}$+$\textit{GRPO} baseline. Results show that V-STAR achieves a {3.87\% relative improvement} in average \texttt{final\_value} and a {5.23\% relative improvement} in maximum \texttt{final\_value}. These gains indicate that value-aware optimized sampling together with Sibling-GRPO enables V-STAR to more reliably identify high-commercial-value recommendation candidates, providing strong support for revenue growth in practical deployments.

\subsection{Ablation Study}
\label{sec:ablation}

We conduct targeted ablation studies to disentangle the contributions of (i) the \emph{decoding strategy} used to construct candidate sets, and (ii) the \emph{training objective} used to optimize the policy.

% \noindent\textbf{Impact of the Decoding Strategy.}
% \label{sec:ablation_decoding}
% To isolate the contribution of decoding strategy, we fix the training way and vary \emph{only} the inference-time candidate construction strategy.
% Specifically, we evaluate the same model of V-STAR on \textit{Industrial} and \textit{Office} using:
% (i) standard beam search,  (ii) top-$K$ sampling (stochastic decoding with truncation to the top-$K$ tokens at each step), and (iii) our Value-Guided Efficient Decoding (VED).\footnote{All decoders operate under the same hard SID-validity constraint. For a fair comparison, we match the candidate budget by returning the same number of final items.}
% Table~\ref{tab:ablation_decoding} summarizes the results. Across both data subsets and all metrics, VED consistently yields the strongest performance.
% Stand beam search concentrates on a small set of high-probability prefixes and may prune low-prior but high-reward branches.
% Top-$K$ sampling increases diversity but is less reliable in ranking quality, as stochastic exploration is not targeted and often spends budget on low-reward branches.
% In contrast, VED delivers consistent improvements by reallocating the same budget to a small number of \emph{decisive} prefixes with high value and high uncertainty, thereby mitigating probability-dominated early pruning and improving reachability of valuable long-tail items.
% Overall, this ablation confirms that value-guided budget allocation is more effective than uniform widening or unguided sampling under the same budget.

\noindent\textbf{Impact of the Decoding Strategy.}\;
\label{sec:ablation_decoding}
To isolate decoding effects, we fix the trained way and vary only the inference-time candidate construction method.
On \textit{Industrial} and \textit{Office}, we compare (i) beam search, (ii) top-K (stochastic sampling from the $K$ highest-probability tokens at each step), and (iii) our VED.\footnote{All methods enforce the same SID-validity constraint and return the same number of final items.}
Table~\ref{tab:ablation_decoding} shows that VED consistently achieves the best performance across datasets and metrics.
Beam search is probability-dominated and may prune low-prior yet high-reward branches, while top-$K$ improves diversity but yields less stable ranking due to untargeted exploration.
In contrast, VED reallocates compute to a small set of decisive prefixes with high value and high uncertainty, mitigating early pruning and improving reachability of valuable long-tail items.

\begin{table}[t]
\centering
\small
\caption{Ablation on decoding strategies.}
\setlength{\tabcolsep}{5pt}
\begin{tabular}{lcccc}
\toprule
Decoding& \multicolumn{2}{c}{Industrial} & \multicolumn{2}{c}{Office} \\
\cmidrule(lr){2-3} \cmidrule(lr){4-5}
 Strategy 
& NDCG@10 $\uparrow$ & HR@10 $\uparrow$
& NDCG@10 $\uparrow$ & HR@10 $\uparrow$ \\
\midrule
Beam Search & 0.1194 & 0.1606 & 0.1299  & 0.1684 \\
Top-K       & 0.1090   & 0.1538  &  0.1226 &  0.1644\\
Ours        & \textbf{0.1217} & \textbf{0.1641} & \textbf{0.1340}  & \textbf{0.1746}  \\
\bottomrule
\end{tabular}
\vspace{-0.7em}
\label{tab:ablation_decoding}
\end{table}

To isolate what drives the gain in VED, we keep the sampling procedure and candidate budget, and vary only the prefix priority score for selecting expansion nodes.
Table~\ref{tab:ablation_acquisition} compares three rules: value-only ($V_\phi$), entropy-only ($\mathcal{H}_\theta$), and the joint score $G(s)$ (Eq.~\ref{eq:acq}) combining $V_\phi$ and $\mathcal{H}_\theta$.
Value-only may over-exploit and miss low-prior yet high-reward branches, whereas entropy-only may over-explore high-uncertainty but low-reward regions.
The joint score is consistently best, indicating that uncertainty mainly serves as an expansion gate: when value is high but confidence is already saturated, further expansion is redundant; uncertainty redirects budget to high-value yet under-resolved prefixes where exploration can still improve reachability.

\begin{table}[t]
\centering
\small
\caption{Ablation of expand rules in VED.}
\setlength{\tabcolsep}{6pt}
\begin{tabular}{l cc cc}
\toprule
Expansion 
& \multicolumn{2}{c}{Industrial} & \multicolumn{2}{c}{Office} \\
\cmidrule(lr){2-3} \cmidrule(lr){4-5}
Rule& NDCG@10 $\uparrow$ & HR@10 $\uparrow$
& NDCG@10 $\uparrow$ & HR@10 $\uparrow$ \\
\midrule
% $\log \pi$   &  &  &  &  \\
$V_\phi(s)$ & 0.1202 & 0.1627  & 0.1333 & 0.1742  \\
$\mathcal{H}_\theta(s)$ & 0.1198 & 0.1604 & 0.1259  & 0.1688   \\
$G(s)$ & \textbf{0.1217} & \textbf{0.1641} & \textbf{0.1340}  & \textbf{0.1746}  \\
\bottomrule
\end{tabular}
%\vspace{2pt}
\label{tab:ablation_acquisition}
\end{table}
%\textbf{Uniform} allocates expansions indiscriminately. \textbf{Value-only} uses $V_\phi$ only. \textbf{Entropy-only} uses $\mathcal{H}_\theta$ only. \textbf{Ours} combines both signals.

\noindent\textbf{Impact of Sibling-GRPO.}\;
\label{sec:ablation_sibling_grpo}
To evaluate the impact of the training objective, we fix VED candidate construction $C(x)$ and compare three RL objectives that differ only in how advantages are normalized on the VED-generated prefix tree: (i) GRPO with global group-based normalization over $C(x)$; (ii) Sibling-GRPO with sibling-relative normalization at shared-prefix branching points; and (iii) GRPO+Sibling-GRPO that jointly optimizes both losses.
Table~\ref{tab:ablation_sibling_grpo} shows that standard GRPO performs worst, as global normalization is dominated by inter-cluster variance and compresses intra-group advantages.
Sibling-GRPO improves performance by aligning updates with the prefix-tree topology, focusing gradients on decisive branching nodes.
Finally, the combined objective achieves the highest NDCG and HR. This suggests that global sequence-level alignment from GRPO and structure-aware, branch-level credit assignment from Sibling-GRPO are mutually reinforcing.

\begin{table}[t]
\centering
\small
\caption{Ablation on training objective.}
\setlength{\tabcolsep}{5pt}
\begin{tabular}{lcccc}
\toprule
Training & \multicolumn{2}{c}{Industrial} & \multicolumn{2}{c}{Office} \\
\cmidrule(lr){2-3} \cmidrule(lr){4-5}
Objective
& NDCG@10 $\uparrow$ & HR@10 $\uparrow$
& NDCG@10 $\uparrow$ & HR@10 $\uparrow$ \\
\midrule
GRPO         & 0.1189 & 0.1598 & 0.1302 & 0.1712 \\
Sibling-GRPO &  0.1204&  {0.1640}&  0.1335 & \textbf{0.1749}  \\
Joint &\textbf{0.1217} & \textbf{0.1641} & \textbf{0.1340}  & 0.1746   \\
\bottomrule
\end{tabular}
\vspace{-0.8pt}
\label{tab:ablation_sibling_grpo}
\end{table}

% \section{Case Study}
\begin{figure}[t]
    \centering
    \includegraphics[width=\linewidth]{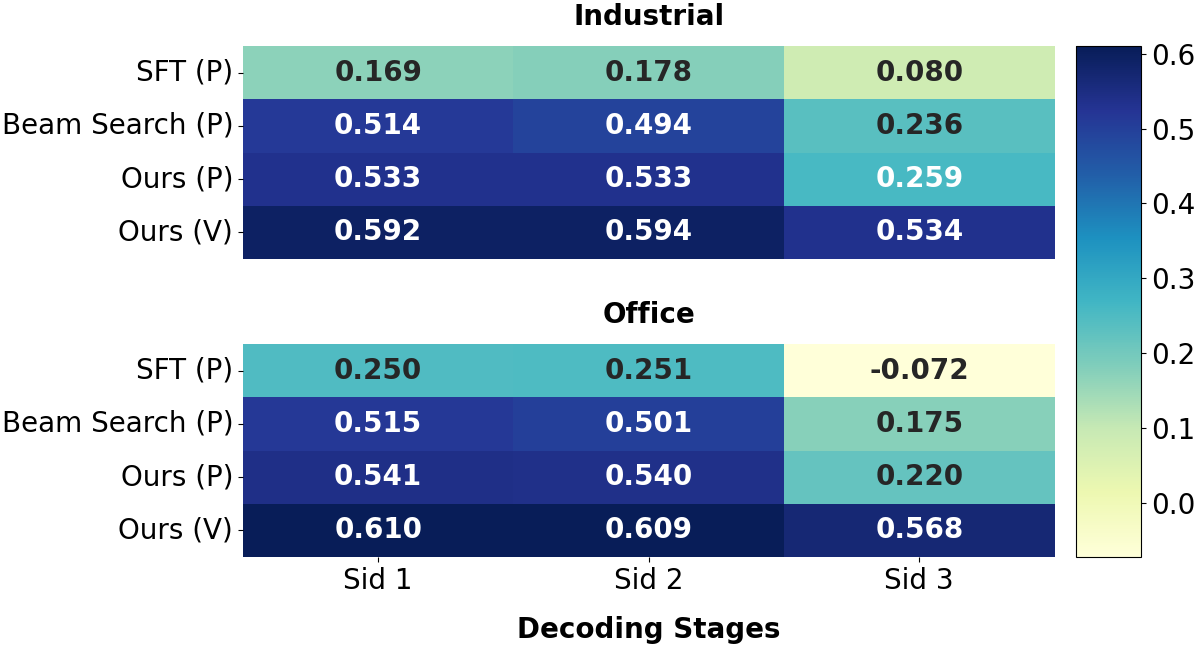}
    \caption{Spearman Correlation $\rho$ of Probability(P) and Value (V) Signals with Ground-truth Rewards.}
    \label{fig:qa_prob}
    % \vspace{-0.4em}
\end{figure}

\begin{figure*}[t]
    \centering
    \includegraphics[width=\linewidth]{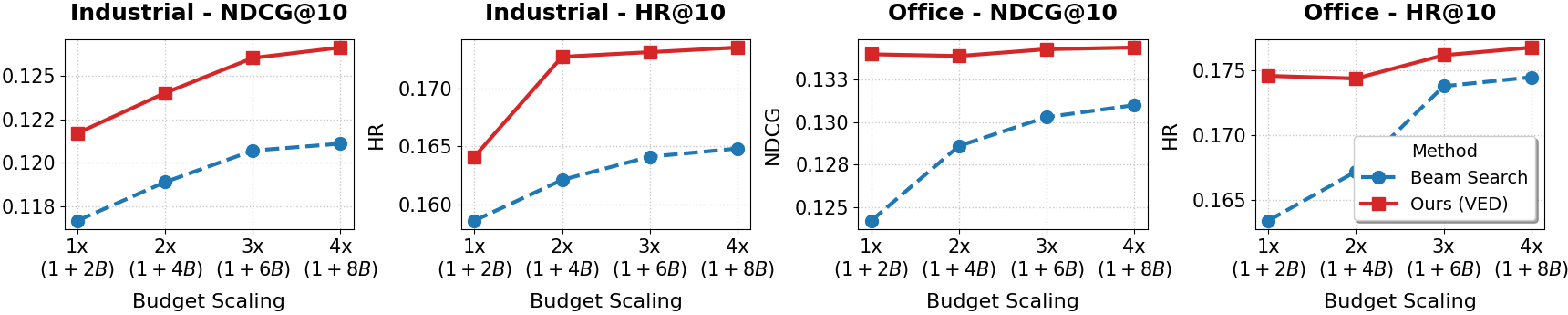}
    \caption{Training time scaling with decoding token budget.}
    \label{fig:scale}
    \vspace{-0.2em}
\end{figure*}

% 定义一个好看的绿色
\definecolor{teal}{HTML}{008080}

% Case Study 2
\begin{table*}[t]
\centering
\small
\setlength{\tabcolsep}{2pt}
\caption{Case Study of different decoding strategies on \textit{Office Products}. SIDs in \textcolor{red}{red} share prefixes with history items. SIDs in \textcolor{teal}{teal} represent novel explorations. SIDs in \textcolor{green}{green} represent ground truth. VED shows higher candidate diversity and successfully discovers the ground-truth (GT) item in a novel branch (\texttt{<a\_20>}), while baselines remain biased toward historical prefixes.}
\label{tab:case_study2_concise}
\begin{tabular}{l p{15.9cm}}
\toprule
\textbf{Context} & \textbf{Key Items \& Candidate SIDs (For non-GT candidates, only the first SID are shown for simplicity)} \\
\midrule
\textbf{History} &
Pencil Sharpener (\texttt{<a\_250>}),
Postcards (\texttt{<a\_225>}),
Bubble Mailer (\texttt{<a\_118>}),
Thank-you Stickers (\texttt{<a\_186>}),
Sharpie Pen (\texttt{<a\_102>}) \\
\midrule
\textbf{GT Item} &
\textbf{ACCUTECK Digital Scale} (\texttt{\textcolor{green}{<a\_20><b\_181><c\_107>}}) \\
\midrule
\midrule
\multirow{2}{*}{\textbf{Beam Search}} &
Stickers (\texttt{\textcolor{red}{<a\_186>}}),
Label Tape (\texttt{\textcolor{teal}{<a\_117>}}),
Sharpie Pen (\texttt{\textcolor{red}{<a\_102>}}),
Postcards (\texttt{\textcolor{red}{<a\_225>}}),
Bubble Mailer (\texttt{\textcolor{red}{<a\_118>}}),
Stickers (\texttt{\textcolor{red}{<a\_186>}}),
Label Printer (\texttt{\textcolor{teal}{<a\_117>}}),
Address Labels (\texttt{\textcolor{red}{<a\_186>}}),
Shipping Labels (\texttt{\textcolor{teal}{<a\_117>}}),
Label Tape (\texttt{\textcolor{teal}{<a\_117>}}) \\
& \textit{(High redundancy: 6/10 items are historical prefixes; GT is missed)} \\
\midrule
\multirow{2}{*}{\textbf{Top-K}} &
Sharpie Pen (\texttt{\textcolor{red}{<a\_102>}}),
Thank-you Labels (\texttt{\textcolor{red}{<a\_186>}}),
Thank-you Stickers (\texttt{\textcolor{red}{<a\_186>}}),
Sharpie Marker (\texttt{\textcolor{red}{<a\_102>}}),
Sharpie Marker (\texttt{\textcolor{red}{<a\_102>}}),
Foam Board (\texttt{\textcolor{teal}{<a\_242>}}),
EXPO Markers (\texttt{\textcolor{teal}{<a\_178>}}),
Bubble Mailer (\texttt{\textcolor{red}{<a\_118>}}),
Glue Stick (\texttt{\textcolor{teal}{<a\_131>}}),
File Folders (\texttt{\textcolor{teal}{<a\_204>}}) \\
& \textit{(Prefix-biased: 6/10 items are historical prefixes; GT is missed)} \\
\midrule
\multirow{2}{*}{\textbf{Ours (VED)}} &
Thank-you Stickers (\texttt{\textcolor{red}{<a\_186>}}),
Thank-you Labels (\texttt{\textcolor{red}{<a\_186>}}),
Sharpie Pen (\texttt{\textcolor{red}{<a\_102>}}),
\textbf{ACCUTECK Digital Scale (GT)} (\texttt{\textcolor{green}{<a\_20><b\_181><c\_107>}}),
Round Labels (\texttt{\textcolor{red}{<a\_186>}}),
File Folders (\texttt{\textcolor{teal}{<a\_187>}}),
Sharpie Marker (\texttt{\textcolor{teal}{<a\_102>}}),
EXPO Markers (\texttt{\textcolor{teal}{<a\_178>}}),
Poly Mailers (\texttt{\textcolor{teal}{<a\_118>}}),
Glue (\texttt{\textcolor{teal}{<a\_242>}}) \\
& \textit{(Finds GT via novel \texttt{<a\_20>} branch while keeping diverse useful neighbors)} \\
\bottomrule
% \vspace{-1em}
\end{tabular}
\end{table*}

\subsection{Further Analysis}
\label{sec:further_analysis}

% This section provides a deeper diagnosis of the probability--reward mismatch and explains why value guidance and structured optimization improve both candidate discovery and learning stability.

% \paragraph{Probability--reward alignment across SID levels.}
% \label{sec:prob_reward_levels}
\noindent\textbf{How well do probability and value align with ground-truth reward across SID levels?} To answer this, we form a candidate pool $\mathcal{C}(x)$ of $64$ SIDs for each query $x$ by temperature sampling from 3 different models: SFT model, RL model trained with beam search+standard GRPO, and our V-STAR model. The temperature is set to 1.5 to cover various probability range. At each level $\ell\in\{1,2,3\}$, we compute Spearman correlation $\rho$ between prefix log-probability $\log\pi_\theta(y_{\leq\ell}\mid x)$ and reward $R(x,y_{\leq\ell})$, where $R(x,y_{\leq\ell})$ is the average reward of candidate in $\mathcal{C}(x)$ that share prefix $y_{\leq\ell}$. For V-STAR, we also compute $\rho$ between value prediction $V_{\phi}(x,y_{\leq\ell})$ and $R(x,y_{\leq\ell})$. We report $\rho$ averaged over queries in Fig.~\ref{fig:qa_prob}. From the figure we can see that: (i) the predicted value $V_{\phi}(x,y_{\leq\ell})$ aligns best with $R(x,y_{\leq\ell})$, (ii) for probability--reward alignment, {V-STAR} $>$ {beam search+GRPO} $>$ {SFT}, indicating that RL improves the probability landscape and V-STAR further enhances such improvement, (iii) the correlation gap is largest at $\ell{=}3$, where semantically similar leaf items reduce discriminability and weaken likelihood--reward correlation, while the value model remains strongly reward-aligned.

% \paragraph{Beam pruning of high-reward candidates.}
% \label{sec:beam_pruning}
% \noindent\textbf{Does value guidance reduce beam pruning of high-reward candidates?}
% Within the temperature-sampled pool $\mathcal{C}(x)$, we define the high-reward set as the top $10\%$ candidates by $R(x,y)$ and report the fraction not covered by beam search (high-reward miss rate), averaged across the $100$ queries.
% Figure~\ref{fig:pruning} summarizes the results.
% Beam pruning is severe on \textit{Industrial} for SFT and remains high for the Beam Search baseline, whereas Ours reduces the miss rate below $1\%$.
% On \textit{Office}, where beam pruning is already small, Ours still yields a consistent reduction.
% Overall, these results suggest that value-guided sampling recovers high-reward regions that are systematically missed by probability-based beams.

\begin{table}[t]
\centering 
\small
\caption{Diversity and max reward of candidate sets.}
\setlength{\tabcolsep}{5pt}
\begin{tabular}{lcccc}
\toprule
\multirow{2}{*}{{Method}}& \multicolumn{2}{c}{Industrial} & \multicolumn{2}{c}{Office} \\
\cmidrule(lr){2-3} \cmidrule(lr){4-5}
 &  $\mathrm{Div}(\mathcal{C}) \uparrow$ & MaxReward $\uparrow$
       &  $\mathrm{Div}(\mathcal{C}) \uparrow$ & MaxReward $\uparrow$ \\
\midrule
Beam search & 0.7949 & 0.2303 & 0.8624 & 0.2321 \\
Top-K       & 0.8089 & 0.2246 & \textbf{0.8856} & 0.2160 \\
Ours (VED)  & \textbf{0.8167} & \textbf{0.2475} & 0.8852 & \textbf{0.2473}  \\
\bottomrule
\end{tabular}
%\vspace{2pt}
\label{tab:cand_div_maxreward}
\end{table}
\noindent\textbf{Does VED increase candidate-set diversity without sacrificing quality?}
We measure candidate-set diversity in SID space by calculating the average pairwise dissimilarity in candidate pool $\mathcal{C}(x)$. For two SIDs $v_i$ and $v_j$ (each with $L$ tokens), we define their similarity as the normalized length of the longest common prefix (LCP), i.e., $\mathrm{Sim}_{\mathrm{SID}}(v_i,v_j)=\mathrm{LCP}(v_i,v_j)/L$. 
Diversity is then computed as the average of $1-\mathrm{Sim}_{\mathrm{SID}}(v_i,v_j)$ over all candidate pairs in $\mathcal{C}(x)$, where a larger value indicates fewer near-duplicate SIDs and broader coverage across SID branches.
To ensure diversity is not obtained by sacrificing strong candidates, we also report the best-in-set reward $\max_{v\in\mathcal{C}} R(v)$, averaged over test queries. Three decoding strategies are tested: beam search, top-K, and our VED.
Table~\ref{tab:cand_div_maxreward} reports results with $|\mathcal{C}(x)|=64$. VED yields the highest (or matched-highest) diversity across datasets while consistently improving the best-in-set reward, indicating reduced SID redundancy and better discovery of high-reward candidates without trading off candidate quality. A detailed case study is given in Table~\ref{tab:case_study2_concise}.

\noindent\textbf{Does VED exhibit more favorable scaling property than beam search under training-time decoding budgets?}
We compare VED with beam search under matched training-time decoding budgets. Specifically, we scale the decoding token budget by the number of model-evaluated tokens during candidate construction (excluding prompt prefill).
For $L{=}3$ with beam width $B=16$, the $1\times$ budget is $1{+}2B$, and we proportionally expand it to $2\times$ ($1{+}4B$), $3\times$ ($1{+}6B$), and $4\times$ ($1{+}8B$). Similarly, VED is constrained to the same budgets at each scale.
Figure~\ref{fig:scale} shows that VED consistently outperforms beam search across budgets and metrics.
On \textit{Industrial}, VED yields larger gains at low budgets ($1\times\!\rightarrow\!2\times$), whereas widening the beam provides smaller increments, indicating higher compute efficiency from value-guided expansion.
On \textit{Office}, VED maintains a clear margin at every scale, and $1\times$ VED often matches or exceeds $3\times$--$4\times$ beam search. This suggests that {value-guided expansion is more compute-efficient} than indiscriminately widening the beam.
Overall, VED converts additional training-time decoding compute into larger improvements than likelihood-driven widening by reducing redundant expansions and prioritizing high-value branches.

\vspace{-1em}

% \section{Conclusion}
\section{Conclusion}
This paper studies candidate construction for Semantic-ID generative recommendation, where probability-driven decoding (e.g., beam search) tends to over-exploit high-probability sibling branches and provides limited reward contrast for RL. We propose V-STAR, which couples Value-Guided Efficient Decoding (VED) for budgeted candidate construction with Sibling-GRPO for structure-aware policy optimization. VED allocates a fixed decoding budget to a small set of high-potential branching points, improving reachability and candidate diversity without exhaustive tree search. Sibling-GRPO leverages the prefix-tree structure to compute sibling-relative learning signals, stabilizing optimization under prefix coupling. Extensive experiments show that V-STAR consistently improvements over strong generative baselines under strict token budgets, and additional analyses verify better value--reward alignment and more effective use of extra decoding compute.

% \section{Acknowledgments}

% Identification of funding sources and other support, and thanks to
% individuals and groups that assisted in the research and the
% preparation of the work should be included in an acknowledgment
% section, which is placed just before the reference section in your
% document.

% This section has a special environment:
% \begin{verbatim}
%   \begin{acks}
%   ...
%   \end{acks}
% \end{verbatim}
% so that the information contained therein can be more easily collected
% during the article metadata extraction phase, and to ensure
% consistency in the spelling of the section heading.

% Authors should not prepare this section as a numbered or unnumbered {\verb|\section|}; please use the ``{\verb|acks|}'' environment.

% \section{Appendices}

% If your work needs an appendix, add it before the
% ``\verb|\end{document}|'' command at the conclusion of your source
% document.

% Start the appendix with the ``\verb|appendix|'' command:
% \begin{verbatim}
%   \appendix
% \end{verbatim}
% and note that in the appendix, sections are lettered, not
% numbered. This document has two appendices, demonstrating the section
% and subsection identification method.

\bibliographystyle{ACM-Reference-Format}
\bibliography{sample-base}

\end{document}